\documentclass[11pt,a4paper]{article}
\usepackage[hyperref]{emnlp2020}
\usepackage{times}
\usepackage{latexsym}

\usepackage{microtype}
\usepackage{graphicx}
\usepackage{amsmath}
\usepackage{multirow}
\usepackage{booktabs}
\aclfinalcopy 

\title{High-order Semantic Role Labeling}

\author{
	Zuchao Li$^{1,2,3}$,
	Hai Zhao$^{1,2,3,}$\thanks{$\ $ Corresponding author. This paper was partially supported by National Key Research and Development Program of China (No. 2017YFB0304100), Key Projects of National Natural Science Foundation of China (U1836222 and 61733011), Huawei-SJTU Long Term AI Project, Cutting-edge Machine Reading Comprehension and Language Model. Rui Wang was partially supported by JSPS grant-in-aid for early-career scientists (19K20354): ``Unsupervised Neural Machine Translation in Universal Scenarios" and NICT tenure-track researcher startup fund ``Toward Intelligent Machine Translation".},
	Rui Wang$^{4}$,
	and Kevin Parnow$^{1,2,3}$
	\\
	$^1$Department of Computer Science and Engineering, Shanghai Jiao Tong University (SJTU)\\
	$^2$Key Laboratory of Shanghai Education Commission for Intelligent Interaction\\
	and Cognitive Engineering, Shanghai Jiao Tong University, Shanghai, China\\
	$^3$MoE Key Lab of Artificial Intelligence, AI Institude, Shanghai Jiao Tong University, China\\
	$^4$National Institute of Information and Communications Technology (NICT), Kyoto, Japan \\
	{\tt \small charlee@sjtu.edu.cn, zhaohai@cs,sjtu.edu.cn, wangrui@nict.go.jp, parnow@sjtu.edu.cn} \\
}

\date{}

\begin{document}
\maketitle
\begin{abstract}
	Semantic role labeling is primarily used to identify predicates, arguments, and their semantic relationships. Due to the limitations of modeling methods and the conditions of pre-identified predicates, previous work has focused on the relationships between predicates and arguments and the correlations between arguments at most, while the correlations between predicates have been neglected for a long time.
	High-order features and structure learning were very common in modeling such correlations before the neural network era. In this paper, we introduce a high-order graph structure for the neural semantic role labeling model, which enables the model to explicitly consider not only the isolated predicate-argument pairs but also the interaction between the predicate-argument pairs.
	Experimental results on 7 languages of the CoNLL-2009 benchmark show that the high-order structural learning techniques are beneficial to the strong performing SRL models and further boost our baseline to achieve new state-of-the-art results.
\end{abstract}

\section{Introduction}

Linguistic parsing seeks the syntactic/semantic relationships between language units, such as words or spans (chunks, phrases, etc.). The algorithms usually use factored representations of graphs to accomplish the target: a set of \emph{nodes} and relational \emph{arcs}. The types of features that the model can exploit in the inference depend on the information included in the factorized \emph{parts}.

Before the introduction of deep neural networks, in syntactic parsing (a kind of linguistic parsing),  several works  \cite{mcdonald-pereira-2006-online, carreras-2007-experiments, koo-collins-2010-efficient, zhang-mcdonald-2012-generalized, ma-zhao-2012-fourth} showed
that high-order parsers utilizing richer factorization information achieve higher accuracy than low-order ones due to the extensive decision history that can lead to significant improvements in inference \cite{chen-etal-2010-improving}.

\begin{figure*}[h]
	\centering
	\includegraphics[width=1.0\textwidth]{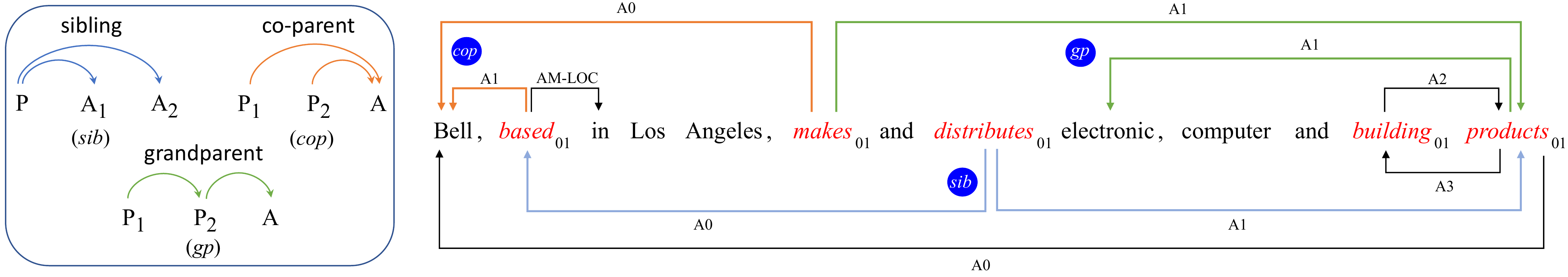} 
	\caption{Left is the second-order parts (structures) considered in this paper, where the $P$ stands for a predicate, and $A$ stands for argument. Right is an example of semantic role labeling from the CoNLL-09 training dataset.}
	\label{fig:example}
\end{figure*}

Semantic role labeling (SRL) \cite{gildea2002automatic,zhao-kit-2008-parsing,zhao-etal-2009-semantic,zhao2013integrative} captures the predicate-argument structure of a given sentence, and it is defined as a shallow semantic parsing task, which is also a typical linguistic parsing task. Recent high-performing SRL models \cite{he-etal-2017-deep,marcheggiani-etal-2017-simple,he-etal-2018-jointly,strubell-etal-2018-linguistically,he-etal-2018-syntax,cai-etal-2018-full}, whether labeling arguments for a single predicate using sequence tagging model at a time or classifying the candidate predicate-argument pairs, are (mainly) belong to first-order parsers. High-order information is an overlooked potential performance enhancer; however, it does suffer from an enormous spatial complexity and an expensive time cost in the inference stage. As a result, most of the previous algorithms for high-order syntactic dependency tree parsing are not directly applicable to neural parsing. In addition, the target of model optimization, the high-order relationship, is very sparse. It is not as convenient for training the model with negative likelihood as the first-order structure is because the efficient gradient backpropagation of parsing errors from the high-order parsing target is indispensable in neural parsing models.

To alleviate the computational and graphic memory occupation challenges of explicit high-order modeling in the training and inference phase, we propose a novel high-order scorer and an approximation high-order decoding layer for the SRL parsing model. For the high-order scorer, we adopt a triaffine attention mechanism, which is extended from the biaffine attention \cite{Dozat2017Deep}, for scoring the second-order parts. In order to ensure the high-order errors backpropagate in the training stage and to output the part score of the first-order and highest-order fusion in the highest-scoring parse search stage during decoding, inspired by \cite{lee-etal-2018-higher,wang-etal-2019-second}, we apply recurrent layers to approximate the high-order decoding iteratively and hence make it differentiable. 

We conduct experiments on popular English and multilingual benchmarks. From the evaluation results on both test and out-of-domain sets, we observe a statistically significant increase in semantic-$F_1$ score with the second-order enhancement and report new state-of-the-art performance in all test set of 7 languages except for the out-of-domain test set in English. Additionally, we also evaluated the results of the setting without pre-identified predicates and compared the effects of every different high-order structure combination on all languages to explore how the high-order structure contributes and how its effect differs from language to language.
Our analysis of the experimental results shows that the explicit higher-order structure learning yields steady improvements over our replicated strong BERT baseline for all scenarios.

\section{High-order Structures in SRL}

High-order features or structure learning is known to improve linguistic parser accuracy. In dependency parsing, high-order dependency features encode more complex sub-parts of a dependency tree structure than the features based on first-order, bigram head-modifier relationships. The clear trend in dependency parsing has shown that the addition of such high-order features improves parse accuracy \cite{mcdonald-pereira-2006-online,carreras-2007-experiments,koo-collins-2010-efficient,zhang-mcdonald-2012-generalized,ma-zhao-2012-fourth}.  We find that this addition can also benefit semantic parsing, as a tree is a specific form of a graph, and the high-order properties that exist in a tree apply to the graphs in semantic parsing tasks as well.

For a long time, SRL has been formulated as a sequential tagging problem or a candidate pair (word pair) classification problem.  In the pattern of sequential tagging, only the arguments of one single predicate are labeled at a time, and a CRF layer is generally considered to model the relationship between the arguments implicitly \cite{zhou-xu-2015-end}. In the candidate pair classification pattern, \citet{he-etal-2018-jointly} propose an end-to-end approach for jointly predicting all predicates, arguments, and their relationships. This pattern focuses on the first-order relationship between predicates and arguments and adopts dynamic programming decoding to enforce the arguments' constraints. From the perspective of existing SRL models, high-order information has long been ignored. 
Although current first-order neural parsers could encode the high-order relationships implicitly under the stacking of the self-attention layers, the advantages of explicit modeling over implicit modeling lie in the lower training cost and better stability. 
This performance improvement finding resultant of high-order features or structure learning suggests that the same benefits might be observed in SRL. Thus, this paper intends to explore the integration and effect of high-order structures learning in the neural SRL model.

The trade-offs between rich high-order structures (features), decoding time complexity, and memory requirements need to be well considered, especially in the current neural models.  The work of  \citet{li2020global} suggests that with the help of deep neural network design and training, exact decoding can be replaced with an approximate decoding algorithm, which can significantly reduce the decoding time complexity at a very small performance loss; however, using high-order structure unavoidably brings problematically high graphic memory demand due to the gradient-based learning methods in the neural network model. Given an input sentence with length $L$, order $J$ of parsing model, the memory required is $O(L^{J+1})$. In the current GPU memory conditions, second-order $J=2$ is the upper limit that can be explored in practice if without pruning. Therefore, we enumerate all three second-order structures as objects of study in SRL, as shown in the left part of Figure \ref{fig:example}, namely \textit{sibling} (sib), \textit{co-parents} (cop), and \textit{grandparent} (gp).

As shown in the SRL example presented in right part of Figure \ref{fig:example}, our second-order SRL model looks at several pairs of arcs: 

	$\bullet$ \textit{sibling} \cite{smith-eisner-2008-dependency, martins-etal-2009-concise}: arguments of the same predicate;
	
	$\bullet$ \textit{co-parents} \cite{martins-almeida-2014-priberam}: predicates sharing the same argument;
	
	$\bullet$ \textit{grandparent} \cite{carreras-2007-experiments}: predicate that is the argument of another predicate.
	

Though some high-order structures have been studied by some related works \cite{yoshikawa-etal-2011-jointly,ouchi-etal-2015-joint,shibata-etal-2016-neural, ouchi-etal-2017-neural,matsubayashi-inui-2018-distance} in Japanese Predicate Argument Structure (PAS) \cite{iida-etal-2007-annotating} analysis and English SRL \cite{yang-zong-2014-multi}, the integration of multiple high-order structures into a single framework and exploring the high-order effects on multiple languages, different high-order structure combinations in a comprehensive way on popular CoNLL-2009 benchmark is the first considered in this paper and thus takes the shape of the main novelties of our work.

\section{Model}

\subsection{Overview}

SRL can be decomposed into four subtasks: predicate identification, predicate disambiguation, argument identification, and argument classification. Since the CoNLL-2009 shared task identified all predicates beforehand, we mainly focus on identifying arguments and labeling them with semantic roles. We formulate the SRL task as a set of arc (and label) assignments between part of the words in the given sentence instead of focusing too much on the roles played by the predicate and argument individually. The predicate-argument structure is regarded as a general dependency relation, with predicate as the \textit{head} and argument as the \textit{dependent} (\textit{dep}) role. Formally, we describe the task with a sequence $X = w_1, w_2, ..., w_n$, a set of unlabeled arcs $Y_{arc} = \mathcal{W} \times \mathcal{W}$, where $\times$ is the cartesian product, and a set of labeled predicate-argument relations $Y_{label} = \mathcal{W} \times \mathcal{W} \times \mathcal{R}$ which, along with the set of arcs, is the target to be predicted by the model. $\mathcal{W}  = \{w_1, w_2, ..., w_n\}$ refers to the set of all words, and $\mathcal{R}$ is the candidate semantic role labels.

\begin{figure}
	\centering
	\includegraphics[width=0.51\textwidth]{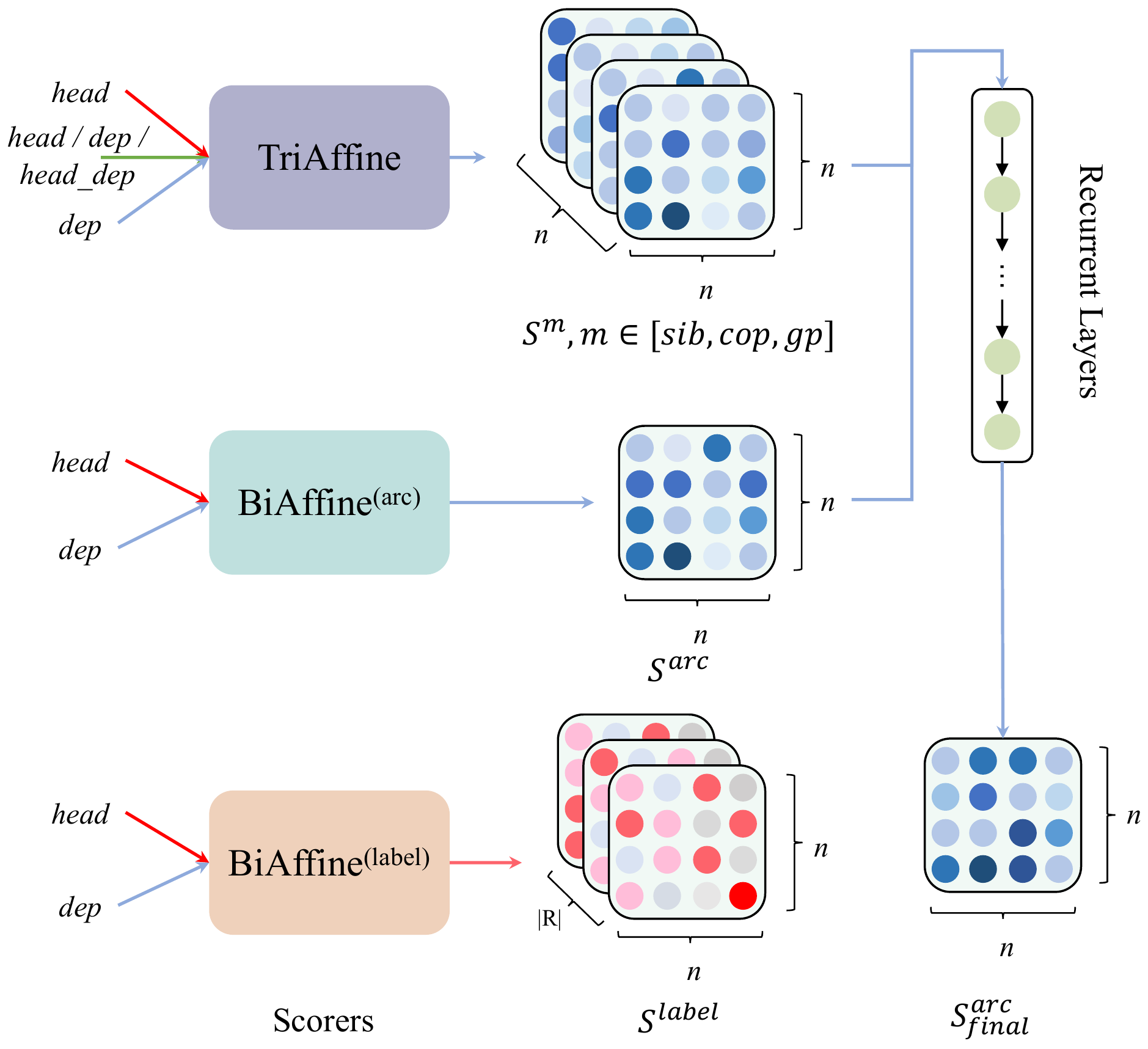} 
	\caption{The proposed model architecture.}
	\label{fig:model}
\end{figure}

Our proposed model architecture for second-order SRL is shown in Figure \ref{fig:model}, which is inspired and extended from \cite{lee-etal-2018-higher,li2019dependency, wang-etal-2019-second}\footnote{Code available at \url{https://github.com/bcmi220/hosrl}.}. The baseline is a first-order SRL model \cite{li2019dependency}, which only considers predicate-argument pairs. Our proposed model composes of three modules: contextualized encoder, scorers, and variational inference layers. Given an input sentence, it first computes contextualized word representations using a BiLSTM encoder on the concatenated embedding. The contextualized word representations are then fed into three scorers to give the arc score, arc label score, and high-order part score following the practice of \citet{Dozat2017Deep}. 
Rather than looking for a model in which exact decoding is tractable, which could be even more stringent for parsing semantic graphs than for dependency trees, we embrace approximate decoding strategies and introduce the variational inference layers to make the high-order error fully differentiable.

\subsection{Encoder}

Our model builds the contextualized representations by using a stacked bidirectional Long Short-term Memory neural network (BiLSTM) \cite{hochreiter1997long} to encode the input sentence. Following \cite{he-etal-2018-syntax,cai-etal-2018-full,li2019dependency}, the input vector is the concatenation of of multiple source embeddings, including a pre-trained word embedding, a random initialized lemma embedding, a predicate indicator embedding, and pre-trained language model layer features; however, unlike their work, we do not use Part-Of-Speech (POS) tag embeddings\footnote{POS tags are also considered to be a kind of syntactic information.}, which enables our model to be truly syntactic-agnostic. Additionally, we use pre-trained language model (PLM) layer features because the latest work \cite{he-etal-2018-syntax,li-etal-2018-unified,li2019dependency,he-etal-2019-syntax} has demonstrated it can boost performance of SRL models. Since these language models were trained at the character- or subword-level, and the out-of-vocabulary (OOV) problem was solved well, we did not use the the bi-directional LSTM-CNN architecture, where convolutional neural networks (CNNs) encode characters inside a word into a character-level representation. Finally, the contextualized representation is obtained as: 
$$H = \textbf{BiLSTM}(E),$$
where $e_i = e_i^{word} \oplus e_i^{lemma} \oplus e_i^{indicator} \oplus e_i^{plm}$ is the concatenation ($\oplus$) of the multiple source embeddings of word $w_i$, $E$ represents $[e_1, e_2, ... , e_n]$, and $H = [h_1, h_2, ... , h_n]$ represents the hidden states (i.e., the contextualized representation) of the BiLSTM encoder.

\subsection{Scorers}

Before scoring the arcs and their corresponding role labels, we adopt two multi-layer perceptron (MLP) layers in different scorers to obtain lower-dimensional and role-specific representations of the encoder outputs to strip away irrelevant information from feature extraction.
\begin{align*}
	h_i^{(u-head)} &= \textbf{MLP}^{(u-head)}(h_i), \\ 
	h_i^{(u-dep)} &= \textbf{MLP}^{(u-dep)}(h_i), \\ 
	\it{u} \in \{\textit{arc}, & \textit{label}\}.
\end{align*}

\paragraph{First-order Arc and Label Scorers:} In order to score the first-order parts (arcs and labels), we adopt the biaffine classifier proposed by \cite{Dozat2017Deep} to compute the possibility of arc existence and label for dependency $i \rightarrow j$ via biaffine attention.
\begin{equation} \label{equation:biaffine}
\textbf{BiAF}(\textbf{v}_i, \textbf{v}_j) =  \left[
\begin{array}{c}
\mathbf{v}_{j}    \\
1
\end{array}
\right]^\mathrm{T}
\mathbf{U}^\textit{1st}  \mathbf{v}_{i}
\end{equation}
\begin{align*}
S^{u}_{i,j} = S^{u}_{i \rightarrow j} = \textbf{BiAF}^{(u)}& (h_i^{(u-head)}, h_j^{(u-dep)}), \\
\it{u} \in \{\textit{arc}, & \textit{label}\},
\end{align*}
where the dimensional size of weight matrix $\mathbf{U}^\textit{1st}$ is $(d+1) \times d$ in the $\textbf{BiAF}^{(arc)}$ function , and $(d+1) \times |\mathcal{R}| \times d$ in the $\textbf{BiAF}^{(label)}$ function, $d$ is the hidden size of the MLPs.

\paragraph{Second-order part scorer:} Inspired by \cite{Dozat2017Deep,wang-etal-2019-second,zhang2020efficient}, we extend the original biaffine attention to a triaffine attention for scoring the second-order parts. Similarly, we employ extra MLPs to perform dimension reduction and feature extraction. Additionally, an extra role $head\_dep$ apart from $head$ and $dep$ is introduced by the \textit{grandparent} parts. This role is both the predicate of an argument and the argument of the other predicate.
\begin{align*}
h_i^{(m-head)} &= \textbf{MLP}^{(m-head)}(h_i), \\
h_i^{(m-dep)} &= \textbf{MLP}^{(m-dep)}(h_i), \\
h_i^{(head\_dep)} &= \textbf{MLP}^{(head\_dep)}(h_i), \\
\it{m} \in \{\textit{sib}, & \textit{cop}, \textit{gp}\}.
\end{align*}
To reduce the computation and memory cost, we only use an arc triaffine function to compute scores of second-order parts; the label triaffine scorer is not considered. A triaffine function is defined as follows:
\begin{equation} \label{equation:triaffine}
\textbf{TriAF}(\mathbf{v}_i, \mathbf{v}_j, \mathbf{v}_k) =
\left[
\begin{array}{c}
\mathbf{v}_{k} \\
1
\end{array}
\right]^\mathrm{T}
{\mathbf{v}_{i}}^\mathrm{T}
\mathbf{U}^\textit{2nd}
\left[
\begin{array}{c}
\mathbf{v}_{j} \\
1
\end{array}
\right]
\end{equation}
\begin{align*}
S^{(sib)}_{i,j,k} = S^{(sib)}_{i \rightarrow j, i \rightarrow k}  =\qquad& \\
\mathbf{TriAF}^{(sib)}(h_i^{(sib-head)}&, h_j^{(sib-dep)}, h_k^{(sib-dep)}), \\
S^{(cop)}_{i,j,k} = S^{(cop)}_{i \rightarrow j, k \rightarrow j}  =\qquad& \\
\mathbf{TriAF}^{(cop)}(h_i^{(cop-head)}&, h_j^{(cop-dep)}, h_k^{(cop-head)}), \\
S^{(gp)}_{i,j,k} = S^{(gp)}_{k \rightarrow i \rightarrow j}  =\qquad\quad& \\
\mathbf{TriAF}^{(gp)}(h_i^{(head\_dep)}&, h_j^{(gp-dep)}, h_k^{(gp-head)}),
\end{align*}
where the weight matrix $\mathbf{U}^\textit{2nd}$ is $(d \times (d+1) \times (d+1))$-dimensional.

\subsection{Variational Inference Layers}

In the first-order model, we adopt the negative likelihood of the golden structure as the loss to train the model,  but in the second-order module of our proposed model, a similar approach will encounter the sparsity problem, as the maximum likelihood estimates cannot be obtained when the number of trainable variables is much larger than the number of observations. In other words, it is not feasible to directly approximate the real distribution with the output distribution of the second-order scorer because of the sparsity of the real distribution.

Computing the arc probabilities based on the first-order and multiple second-order scores outputs can be seen as doing posterior inference on a Conditional Random Field (CRF). As exact inference on this CRF is intractable\cite{wang-etal-2019-second}, we resort to using the variational inference algorithms that allow the model to condition on high-order structures while being fully differentiable. 

The variational inference computes the posterior distribution of unobserved variables in the probability graph model. Then, parameter learning is carried out with the observed variables and the predicted unobservable variables. Mean field variational inference approximates a true posterior distribution with a factorized variational distribution and tries to iteratively minimize its KL divergence. Thus, we use mean field variational inference approximates to obtain the final arc distribution. This inference involves $T$ iterations of updating arc probabilities, denoted as $Q_{i,j}^{(t)}$ for the probabilities of arc $i \rightarrow j$ at iteration $t$. The iterative update process is described as follows:
\begin{align*}
\mathcal{G}_{i,j}^{(t-1)} = \sum_{k \neq i,j} \lbrace Q_{i,k}^{(t-1)}  S^{(sib)}_{i \rightarrow j, i \rightarrow k} + Q_{k,j}^{(t-1)} S^{(cop)}_{i \rightarrow j, k \rightarrow j} \\
+ Q_{k,i}^{(t-1)} S^{(gp)}_{k \rightarrow i \rightarrow j} + Q_{j,k}^{(t-1)} S^{(gp)}_{i \rightarrow j \rightarrow k}
\rbrace,
\end{align*}
\begin{align*}
Q_{i,j}^{(t)} =&
\begin{cases}
\exp(S^{arc}_{i \rightarrow j}  + \mathcal{G}_{i,j}^{(t-1)}), &\text{Arc $i \rightarrow j$ exist}\\
1, &\text{Otherwise}
\end{cases}
\end{align*}
where $\mathcal{G}_{i,j}^{(t-1)} $ is the second-order voting scores, $Q_{i,k}^{(0)} = \textbf{softmax}(S^{arc}_{i,j})$, and $t$ is the updating step.

\citet{zheng2015conditional} stated that multiple mean-field update iterations can be implemented by stacking Recurrent Neural Network (RNN) layers, as each iteration takes $Q$ value estimates from the previous iteration and the unary values (first-order scores) in their original form. In this RNN structure, CRF-RNN, the model parameters therefore can be optimized from the second-order error using the standard backpropagation through time algorithm\cite{rumelhart1985learning,mozer1995focused}. Notably, the number of stacked layers is equal to the iteration steps $T$. Since when $T>5$, increasing the number of iterations usually does not significantly improve results \cite{krahenbuhl2011efficient}, training does not suffer from the vanishing and exploding gradient problem inherent to deep RNNs, and this allows us to use a plain RNN architecture instead of more sophisticated architectures such as LSTMs.

\begin{table}[t!]
	\small
	\centering
	\setlength{\tabcolsep}{4pt}
	\begin{tabular}{lccc}  
		\toprule  
		System&  Pre-training &WSJ &Brown \\
		\midrule
		\multicolumn{4}{l}{\textbf{\textit{w/ pre-identified predicate}}}	\\
		\citet{cai-etal-2018-full} & &89.60 &79.00 \\ 
		\citet{kasai-etal-2019-syntax}$^*$ &  & 88.60 & 77.60\\  
		\citet{zhou2019parsing}$^\dag$ & & 89.28&  \bf 82.82\\ 
		\citet{he-etal-2019-syntax}$^*$ &  & 89.96   & - \cr 
		\bf Ours &   & \bf 90.26 &   80.63 \\
		\midrule
		\citet{he-etal-2018-syntax}$^*$& +E &89.50 & 79.30\\ 
		\citet{li2019dependency}  & +E &90.40 & 81.50\\ 
		\citet{kasai-etal-2019-syntax}$^*$ & +E &90.20 & 80.80\\  
		\citet{lyu-etal-2019-semantic} & +E &90.99 & 82.18 \\ 
		\citet{chen-etal-2019-capturing} & +E &91.06 & 82.72 \\  
		\citet{cai-lapata-2019-semi}$^\dag$ & +E &91.20  & 82.50\\ 
		\bf Ours & +E & \bf 91.44 &  \bf 83.28 \\
		\midrule
		\citet{zhou2019parsing}$^\dag$  & +B &91.20 &\bf 85.87\\ 
		\bf Ours & +B & \bf 91.77 &   85.13 \\
		\midrule
		\midrule 
		\multicolumn{4}{l}{\textbf{\textit{w/o pre-identified predicate}}} \\
		\citet{cai-etal-2018-full} & &85.00 & 72.50\\ 
		\citet{li2019dependency}  & &85.10 & -\\ 
		\citet{zhou2019parsing}$^\dag$ & & 85.86&  \bf 77.47\\ 
		\bf Ours &   & \bf 86.16 &   74.20 \\
		\midrule
		\citet{he-etal-2018-syntax}$^*$ & +E & 83.30 & -\\ 
		\citet{li2019dependency} & +E &85.30 &  74.20\\ 
		\bf Ours &  +E & \bf  87.12 &   \bf 76.65 \\
		\midrule
		\citet{zhou2019parsing}$^\dag$ & +B &88.17& \bf 81.58\\ 
		\bf Ours & +B & \bf 88.70 &  80.29\\
		\bottomrule  
	\end{tabular}
	\caption{Semantic-$F_1$score on CoNLL-2009 English treebanks. WSJ is used for evaluating the in-domain performance and Brown for the out-of-domain. ``$^*$" denotes that the model uses syntactic information for enhancement, and ``$^\dag$" represents the model is trained with other tasks jointly. ``+E" stands for using ELMo as pre-trained PLM features, ``+B" for using BERT.}\label{english_results}
\end{table}

\begin{table*}[t!]
	\centering
	\scalebox{0.9}{
		\begin{tabular}{lcccccccc}
			\toprule
			System & CA  & CS & DE & EN & ES &  JA & ZH & Avg. \\
			\midrule
			\multicolumn{9}{l}{\textbf{\textit{w/ pre-identified predicate}}}	\\
			CoNLL-2009 ST  & 80.3  & 85.4 & 79.7 & 85.6  & 80.5 & 78.2  & 78.6 & 81.19 \\
			\citet{zhao-etal-2009-multilingual-dependency} & 80.3  & 85.2 & 76.0 & 86.2  & 80.5 & 78.2  & 77.7 & 80.59 \\
			\citet{roth-lapata-2016-neural} & $-$  & $-$ & 80.1 & 87.7 & 80.2 & $-$ & 79.4 &  $-$ \\
			\citet{marcheggiani-etal-2017-simple} & $-$  & 86.0 & $-$ & 87.7  & 80.3 & $-$ & 81.2 &  $-$ \\
			\citet{mulcaire-etal-2018-polyglot} & 79.45 & 85.14 & 69.97 & 87.24 & 77.32 & 76.00 & 81.89 & 79.57\\
			\citet{kasai-etal-2019-syntax}$^{\text{+E}}$ & $-$  & $-$ & $-$& 90.2 & 83.0 & $-$ & $-$ &  $-$ \\
			\citet{lyu-etal-2019-semantic}$^{\text{+E}}$ & 80.91  & 87.62 & 75.87  & 90.99 & 80.53 & 82.54  & 83.31 & 83.11\\
			\citet{cai-lapata-2019-semi} &  $-$  &  $-$ & 83.80 & 91.20 & 82.90 &  $-$  & 85.00 &  $-$ \\
			\citet{he-etal-2019-syntax} & 84.35  & 88.76 & 78.54 & 89.96  & 83.70 & 83.12  & 84.55 & 84.71 \\
			\citet{he-etal-2019-syntax}$^{\text{+B}}$ & 85.14  & 89.66 & 80.87 & 90.86   & 84.60 & 83.76 & 86.42 & 85.90 \\
			\midrule
			Our baseline & 84.96 & 90.18 & 76.02 & 89.61 & 83.77 &  82.65 &  85.73 & 84.70 \\
			\qquad\quad+HO & 85.37 & 90.60  &  76.41 & 90.26  & 84.39  & 83.25  & 86.02 & 85.19 \\
			Our baseline$^{\text{+B}}$ & 86.40 & 91.48 & 85.21 & 91.23  & 86.60 & 85.55  & 88.24  & 87.82 \\
			\qquad\quad+HO$^{\text{+B}}$ & \bf 86.90  & \bf 91.93 & \bf 85.54 & \bf 91.77  & \bf 86.96 & \bf 85.90  & \bf 88.69 & \bf 88.24\\
			\midrule
			\multicolumn{9}{l}{\textbf{\textit{w/o pre-identified predicate}}}	\\
			Our baseline &  83.69 & 89.22 & 60.06 & 85.71 & 82.54 & 73.68 & 81.46 & 79.48 \\
			\qquad\quad+HO & 84.07  &  89.45  &  60.48 & 86.16 &  83.11 &  74.20  & 82.01 & 79.93\\
			Our baseline$^{\text{+B}}$ & 85.12 & 90.72 & 66.70 & 88.05 & 85.50 &  77.94 &  85.38 & 82.77 \\
			\qquad\quad+HO$^{\text{+B}}$ &  \bf 85.82  &  \bf 91.22 & \bf  67.15 &  \bf 88.70 & \bf 86.00 & \bf 78.88  & \bf 85.68 & \bf 83.35 \\
			\bottomrule
		\end{tabular}
	}
	\caption{Semantic-F$_1$ score on the CoNLL-2009 in-domain test set. The first row is the best result of the CoNLL-2009 shared task \cite{hajic-etal-2009-conll}. ``$^{\text{+E}}$" indicates the model leverages pre-trained ELMo features (only for English), ``$^{\text{+B}}$" indicates the model leverages BERT for all languages.}
	\label{tab:all-results}
\end{table*}

\subsection{Training Objective}
The full model is trained to learn the conditional distribution $P_\theta(\hat{Y}|X)$ of predicted graph $\hat{Y}$ with gold parse graph $Y^*$. Since the parse graph can be factorized to arcs and corresponding labels, the conditional distribution $P_\theta(\hat{Y}|X)$ is also factorized to $P_\theta(\hat{Y}^{(arc)}|X)$ and $P_\theta(\hat{Y}^{(label)}|X)$, given by:
$$P_\theta(\hat{Y}^{(arc)}|X) = \prod_{1 \leq i \leq n,  1 \leq j \leq n} \mathbf{softmax}(Q_{i,j}^{(T)}),$$
$$P_\theta(\hat{Y}^{(label)}|X) = \prod_{1 \leq i \leq n,  1 \leq j \leq n}  \mathbf{softmax}(S_{i,j}^{(label)}).$$
where $\theta$ represents the model parameters. The losses to optimize the model are implemented as cross-entropy loss using negative likelihood to the golden parse:
$$\mathcal{L}^{(arc)}(\theta) = - \sum_{1 \leq i \leq n,  1 \leq j \leq n} \log P( Y^{*(arc)}_{i,j} |X),$$
$$\mathcal{L}^{(label)}(\theta) = - \sum_{(i, j, r) \in Y^{*}} \log P( \langle i \rightarrow j, r \rangle |X) ),$$
where $r \in \mathcal{R}$ is the semantic role label of arc (predicate-argument) $i \rightarrow j$. The final loss is the weighted average of the arc loss $\mathcal{L}^{(arc)}(\theta)$ and the label loss $\mathcal{L}^{(label)}(\theta)$:
$$\mathcal{L}^{(final)}(\theta) = \lambda \mathcal{L}^{(arc)}(\theta) + (1 - \lambda) \mathcal{L}^{(label)}(\theta),$$
where $\lambda$ is the balance hyper-parameter.

\section{Experiments}
\subsection{Setup}
We conduct experiments and evaluate our model on the CoNLL-2009 \cite{hajic-etal-2009-conll} benchmark datasets including 7 languages: Catalan (CA), Czech (CS), German (DE), English (EN), Spanish (ES),  Japanese (JA), and Chinese (ZH). To better compare with previous works, and to bring the model closer to a real-world usage scenario, we consider two SRL setups on all 7 languages: \textit{w/ pre-identified predicate} and \textit{w/o pre-identified predicate}. 
In order to compare with most previous models, the former setup follows official requirements and has predicates identified beforehand in the corpora.
The latter one is consistent with a real scenario; where the model is required to predict all the predicates and their arguments and is therefore relatively more difficult. Since the predicates need to be predicted in the \textit{w/o pre-identified predicate} setup, we treat the identification and disambiguation of predicates as one sequence tagging task, and we adopt \textbf{BiLSTM+MLP} and \textbf{BERT+MLP} sequence tagging architectures to adapt to different requirements.
We directly adopt most hyper-parameter settings and training strategy from \cite{Dozat2017Deep,wang-etal-2019-second}. Please refer to Appendix \ref{hyperparams} for details. 

\subsection{Results And Analysis}

\paragraph{Main Results\footnote{Due to the limited space, we only analyzed the main results. Please refer to Appendix \ref{detailed_results} for detailed results.}} Table \ref{english_results} presents the results on the standard English test set, WSJ (in-domain) and Brown (out-of-domain). 
For a fair comparison with previous works, we report three cases: not using pre-training, using ELMo \cite{peters-etal-2018-deep}, and using BERT \cite{devlin-etal-2019-bert}. Our single model achieves the best performance on the in-domain test set without syntactic information and extra resources for both types of setup, \textit{w/} and \textit{w/o pre-identified predicate}. On the out-of-domain test set, even though \citet{zhou2019parsing} obtains the highest score, their model is joint and likely achieves domain adaptation due to external tasks and resources.
In general, our model achieves significant performance improvements in both in-domain and out-of-domain settings, especially while using pre-training out-of-domain.
Furthermore, the results of using ELMo and BERT show that the stronger pre-training model brings greater improvement.

\begin{table}[t]
	\small
	\centering
	\begin{tabular}{lccc}  
		\toprule  
		System & P & R & $F_1$ \\
		\midrule
		\multicolumn{4}{c}{\textbf{German}}  \\
		\citet{zhao-etal-2009-multilingual-dependency} & - & - &  67.78 \\
		\citet{lyu-etal-2019-semantic} & - & - & 65.69 \\
		Our baseline &  71.34 & 67.73 & 69.49 \\
		\qquad\quad+HO & 71.66 & 69.36 & \bf 70.49 \\
		Our baseline$^{\text{+B}}$ &  71.77 & 70.02 & 70.88 \\
		\qquad\quad+HO$^{\text{+B}}$ &  72.14 & 71.86 & \bf 72.00 \\
		\midrule
		\multicolumn{4}{c}{\textbf{Czech}}  \\
		\citet{zhao-etal-2009-multilingual-dependency} & - & - & 82.66 \\
		\citet{marcheggiani-etal-2017-simple} & 88.00 & 86.50 & 87.20 \\
		\citet{lyu-etal-2019-semantic} & - & - & 86.04 \\
		Our baseline &  91.22 & 89.88 & 90.54 \\
		\qquad\quad+HO & 91.50 & 90.03 & \bf 90.75 \\
		Our baseline$^{\text{+B}}$ &  91.98 & 91.23 & 91.60 \\
		\qquad\quad+HO$^{\text{+B}}$ &  91.87 & 91.61 & \bf 91.74 \\
		\bottomrule  
	\end{tabular}
	\caption{Precision, Recall, and Semantic-$F1$ scores on German and Czech out-of-domain test sets.}\label{other_ood_results}
\end{table}

\paragraph{Multilingual Results} Table \ref{tab:all-results} summarizes the results on CoNLL-2009 standard in-domain test sets of all 7 languages.
The bold results in Table \ref{tab:all-results} are obtained by averaging the every results from 5 training rounds with different random seeds to avoid random initialization impact on the model.
We compare our baseline and full model with previous multilingual works. The performance of our baseline is similar to the model of \citet{he-etal-2019-syntax}, which integrated syntactic information and achieved the best results. This shows that our baseline is a very strong SRL model, 
and owes its success to directly modeling on the full semantic graph rather than separately based on predicates.
Moreover, our model with the proposed high-order structure learning (+ HO) obtains absolute improvements of 0.49\% and 0.42\% F1 without pre-training and with BERT, respectively, achieving the new best results on all benchmarks.
Because the quantities of high-order structures are different among different languages, consistent improvement on 7 languages already shows that our empirical results are convincing.

In addition,  we also report the results of the \textit{w/o pre-identified predicate} setup for all languages, which is a more realistic scenario. The overall decline without pre-identified predicates shows that predicate recognition has a great impact. Especially for German, the obvious drop is probably because the ratio of predicates in the German evaluation set is relatively small and is sensitive to the model parameters; however, in this setup,  our high-order structure learning leads to consistent improvements in all languages with the \textit{w/ pre-identified predicate} setup, demonstrating the effectiveness of the proposed method.

To show the statistical significance of our results, in addition to adopting the above-mentioned common practice in SRL at model-level that reports the average results with multiple runs and random seeds, we further follow the practice in machine translation \cite{koehn-2004-statistical} to conduct a significant test at example level. We sampled the prediction results for 500 times, 50 sentences each time, and evaluated the sampled subset. The result of +HO is significantly higher than that of the baseline model ($p<0.01$), verifying the significance of the results.

\paragraph{Out-of-domain Results}  Besides English, there are also out-of-domain test sets for German and Czech. To verify the generalization capability of our model, we further conduct experiments on these test sets under \textit{w/ pre-identified predicates} and compare results with existing work (in Table \ref{other_ood_results}).  Our model achieves new state-of-the-art results of 70.49\% (German) and 90.75\% (Czech) F1-score, significantly outperforming the previous best system \cite{lyu-etal-2019-semantic}. Furthermore, there is even a gain with using pre-trained BERT, showing that BERT can improve the generalization ability of the model. In addition, we observe that the model (+HO) yields stable performance improvement in recall, which shows the proposed high-order structure learning is beneficial to identifying arguments.

\begin{figure}
	\centering
	\includegraphics[width=0.48\textwidth]{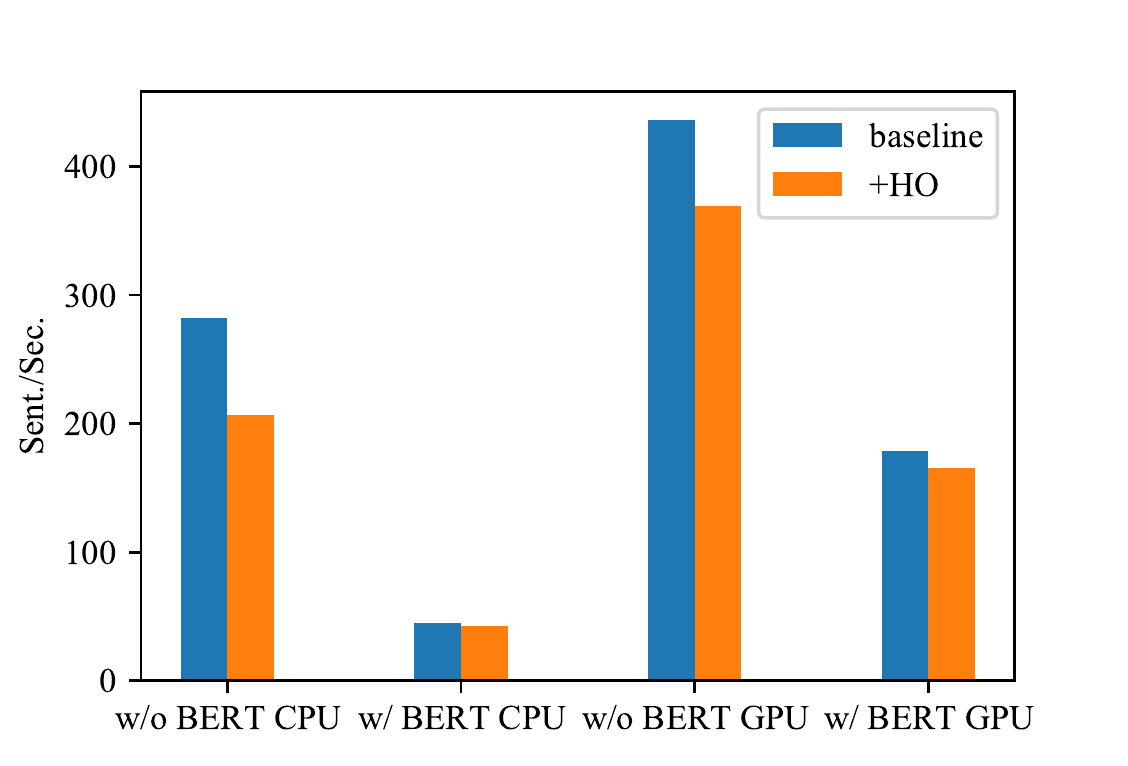} 
	\caption{Parsing speed measured on CoNLL-2009 English test set.}
	\label{fig:speed}
\end{figure}

\paragraph{Time Complexity and Parsing Speed} The time complexity and parsing speed of high-order models have always been concerns. In our proposed high-order model, the time complexity comes from two parts: one is the matrix operations in the biaffine and triaffine attention ($O(d_{\text{BiAF}}^2)$ and $O(d_{\text{TriAF}}^3)$, respectively), where $d_{\text{BiAF}}$ and $d_{\text{TriAF}}$ is the hidden size of the scorer, and the other is the inference procedure ($O(n^3)$), making the total time complexity $O(d_{\text{BiAF}}^2 + d_{\text{TriAF}}^3+ n^3)$, while, for  for our baseline, the full time complexity is $O(d_{\text{BiAF}}^2 + n^2)$. Additionally, in the case of leveraging pre-trained PLM features, the time complexity of encoders such as BERT is a part that cannot be ignored. We measured the parsing speed of our baseline and high-order models on the English test set, both with BERT pre-training and without, on the CPU and GPU, respectively, with an Intel Xeon 6150 CPU and a Titan V100 GPU. The comparison is shown in Figure \ref{fig:speed}. Results show that the speed loss of +HO is 26.7\%, 5.5\%, 15.4\% and 7.6\% in the respective four scenarios, while the speed loss brought by BERT is 84.1\%, 79.5\% on CPU and 60.0\% and 55.2\% on GPU. Therefore, +HO brings a loss of speed, but with GPU acceleration, the loss ratio is reduced. In the case of BERT pre-training, +HO is no longer the bottleneck of parsing speed.

\paragraph{High-order Structures Contribution} To explore the contribution of high-order structures in depth , we consider all possible combinations of structures and conduct experiments on the English test set under the \textit{w/ pre-identified predicate} setup. Table \ref{ablation_study} shows the results of two baseline models (with and without BERT pre-training). Using these structures separately improves our model, as shown in +\textit{sib} with a 0.36 $F_1$ gain; however, the further improvement of applying two structures is limited. For example, model (+\textit{sib}) performs even better than (+\textit{sib}+\textit{gp}). The reason might be that the \textit{sib} (between the arguments) and the \textit{gp} (between the predicates) are two irrelevant structures. Regardless, we can observe that +\textbf{ALL} (the combination of all three structures) model achieves the best performance ( up to 0.65 $F_1$). One possible reason for the result is that the \textit{cop} (between arguments and predicates) sets up a bridge for \textit{sib} and \textit{gp} structures. In other words,  these observations suggest that the three structure learning may be complementary.

We further explored the sources of the higher order structure's improvement in SRL performance. we split the test set into two parts, one with the high-order relationship (cop and gp), and the other without. Taking the CoNLL09 English test set as an example, the total size of the test set is 2399 sentences, and there are 1936 sentences with high-order relationships. We recalculated Sem-F1 for these two subsets, and found that the scores on the subsets with higher-order relationships are significantly higher than those without(>0.4\% F-score). It shows that our model does improve the prediction of high-order structure, rather than a specific type of semantic role. For simple sentences (without HO), the baseline can already parse it very well, which also explains the reason why the improvement in some languages is not great.

\begin{table}[t]
	\small
	\centering
	\setlength{\tabcolsep}{4pt}
	\resizebox{\linewidth}{!}{
	\begin{tabular}{lcccc}  
		\toprule  
		\multirow{2}{*}{System}
		&\multicolumn{2}{c}{w/o BERT}
		&\multicolumn{2}{c}{w/ BERT}
		\\
		\cmidrule(lr){2-3} 
		\cmidrule(lr){4-5} 
		&P / R&F$_1$&P / R &F$_1$\\
		\midrule
		\textbf{baseline} & 91.29 / 88.00 & 89.61  & 92.31 /  90.18 &  91.23 \\
		+\textit{sib} & 91.46 / 88.53 & 89.97 & 92.49 / 90.58 &  91.53 \\
		+\textit{cop} & 91.40 / 88.45 & 89.90 & 92.21 / 90.50 & 91.35 \\
		+\textit{gp} & 91.41 / 88.14 & 89.75 & 92.77 / 90.09 & 91.41 \\
		+\textit{sib}+\textit{cop} &  91.33 / 88.74 & 90.02 & 92.36 / 90.84 & 91.60 \\
		+\textit{sib}+\textit{gp} & 91.21 / 88.58 & 89.87 & 92.64 / 90.44 & 91.53 \\
		+\textit{cop}+\textit{gp} & 91.26 / 88.68  & 89.95 & 92.37 / 90.35 & 91.35 \\
		+\textbf{ALL} & 91.60 / 88.95 & 90.26 & 92.59 / 90.98 & 91.77 \\
		\bottomrule  
	\end{tabular}
	}
	\caption{Effect of different second-order structures and their combination on model performance.}\label{ablation_study}
\end{table}

\section{Related Work}

The CoNLL-2009 shared task advocated performing SRL for multiple languages to promote multilingual NLP applications. 
\cite{zhao-etal-2009-multilingual-dependency} proposed an integrated approach by exploiting large-scale feature sets, while \cite{bjorkelund-etal-2009-multilingual} used a generic feature selection procedure, which yielded significant gains in the multilingual SRL shared task.
With the development of deep neural networks \cite{li-etal-2018-seq2seq,xiao-etal-2019-lattice,zhou-zhao-2019-head,zhang2019effective,zhang2019explicit,li-etal-2019-sjtu,luo2020hierarchical,li2019global,zhang2019semantics} for NLP,
most subsequent SRL works have focused on improving the performance of English, with occasional comparisons to other languages \cite{lei-etal-2015-high,swayamdipta-etal-2016-greedy,roth-lapata-2016-neural,marcheggiani-etal-2017-simple,he-etal-2018-syntax,li-etal-2018-unified,cai-etal-2018-full}.  \citet{mulcaire-etal-2018-polyglot} built a polyglot semantic role labeling system by combining resources from all languages in the CoNLL-2009 shared task for exploiting the similarities between semantic structures across languages. This approach, while convenient, is still far less effective than separate model training on different languages. \citet{lyu-etal-2019-semantic} modeled interactions between argument labeling decisions with a structure refinement network, resulting in an effective model, and outperforming strong factorized baseline models on all 7 languages. \citet{he-etal-2019-syntax} boosted multilingual SRL performance with special focus on the impact of syntax and contextualized word representations and achieved new state-of-the-art results on the CoNLL-2009 benchmarks of all languages, resulting in an effective model and outperforming strong factorized baseline models on all 7 languages

High-order parsing is one of the research hotspots in which first-order parsers meet performance bottlenecks; this has been extensively studied in the literature of syntactic dependency parsing\cite{mcdonald-etal-2005-online,mcdonald-pereira-2006-online,carreras-2007-experiments,koo-collins-2010-efficient,martins-etal-2011-dual,ma-zhao-2012-fourth,gormley-etal-2015-approximation,zhang2020efficient}. In semantic parsing, 
\citet{martins-almeida-2014-priberam} proposed a way to encode high-order parts with hand-crafted features and introduced a novel co-parent part for semantic dependency parsing. \citet{cao-etal-2017-quasi} proposed a quasi-second-order semantic dependency parser with dynamic programming. \citet{wang-etal-2019-second} trained a second-order parser in an end-to-end manner with the help of mean field variational inference and loopy belief propagation approximation. In SRL or related research field, there is also some related work on the improvement of performance by high-order structural information. On the Japanese NAIST Predict-Argument Structure (PAS) dataset, some works \cite{yoshikawa-etal-2011-jointly,ouchi-etal-2015-joint,iida-etal-2015-intra,shibata-etal-2016-neural,ouchi-etal-2017-neural,matsubayashi-inui-2018-distance} mainly studied the relationship between multiple predicates separately, that is, the \textit{gp} and \textit{cp} high-order relationship mentioned in our paper. \cite{yang-zong-2014-multi} considered the interactions between predicate-argument pairs on Chinese Propbank dataset. Although the motivation is consistent with our work, we first consider multiple high-order relationships at the same time within a more uniform framework on more popular benchmarks and for more languages.

\section{Conclusion and Future Work}

In this work, we propose high-order structure learning for dependency semantic role labeling. The proposed framework explicitly models high-order graph structures on a strong first-order baseline model while scoring the correlation of predicted predicate-argument pairs. The resulting model achieves state-of-the-art results on all 7 languages in the CoNLL-2009 test data sets except the out-of-domain benchmark in English. In addition, we consider both given and not-given predicates on all languages, explore the impact of every high-order structure combinations on performance for all languages, and reveal the adaptive range of high-order structure learning on different languages. In future work, we will continue to explore higher-order structures and pruning strategies to reduce the time complexity and memory occupation. 

\bibliographystyle{acl_natbib}
\bibliography{emnlp2020}

\clearpage
\appendix

\section{Appendices}

\subsection{Hyper-parameters and Training Details}\label{hyperparams}

In our experiments, the \textbf{BiLSTM+MLP} predicate tagging model only takes words and lemmas as input, and its encoder structure is the same as our main model, so the hyper-parameters are also consistent with our main model. With the \textbf{BERT+MLP} predicate tagging model, the motivation for choosing this instead of using BERT as embedding in the \textbf{BiLSTM+MLP} architecture is to achieve fair comparability with the results of \cite{zhou2019parsing}.

For the hyper-parameters of our main model, we borrowed most parameter settings from \cite{Dozat2017Deep, wang-etal-2019-second}, including dropout and initialization strategies. Hyper-parameters for our baseline and proposed high-order model are shown in Table \ref{tab:hyper_both}. 
We use 100-dimensional Glove \cite{pennington-etal-2014-glove} pre-trained word embeddings for English and 300-dimensional FastText embeddings \cite{bojanowski-etal-2017-enriching,grave-etal-2018-learning} for all other languages. 
As for the pre-training, ELMo\cite{peters-etal-2018-deep} is only used in English, we take the weighted sum of the 3 layers as the final features, while different versions of  BERT\cite{devlin-etal-2019-bert}  are used in different languages, as shown in Table \ref{tab:bert_versions}, we always use the second-to-last layer outputs as the pre-trained features.

Following the work of \cite{wang-etal-2019-second}, during model training, the training strategy includes two phases. In the first phase, we used Adam \cite{kingma2014adam} and annealed the learning rate $0.5$ every 10,000 steps. When the training reaches 5,000 steps without improvement, the model optimization enters the second phase; the Adam optimizer is replaced by AMSGrad \cite{j.2018on}. We trained the model for maximum 100K update steps with batch sizes of \{4K, 2K, 3K, 4K, 6K, 6K, 6K\} tokens for  CA,  CS,  DE,  EN,  ES,   JA, and ZH, respectively. The training is terminated with an early stopping mechanism when there is no improvement after 10,000 steps on the development sets.

\begin{table}[h]
	\small
	\begin{center}
		\begin{tabular}{lr}
			\hline \hline
			\textbf{Hidden Layer} & \textbf{Hidden Sizes}\\ \hline
			Word Embedding & 100 (en) / 300 (others)\\
			Lemma Embedding & 100 \\
			Predicate Indicator / Sense Emb & 50 / 50 \\
			ELMo/BERT Linear & 100 \\
			Stacked BiLSTM & 3 $\times$ 600 \\
			Biaffine Arc/Label Scorer & 600\\
			Triaffine Arc Scorer & 150\\
			\hline \textbf{Dropouts} & \textbf{Dropout Prob.}\\ \hline
			Word/Lemma/Predicate & 20\%\\
			BiLSTM (FF/recur) & 45\%/25\%\\
			Biaffine Arc/Label Scorer & 25\%/33\%\\
			Triaffine Arc Scorer & 25\%\\
			\hline \textbf{Optimizer \& Loss} & \textbf{Value}\\ \hline
			Balance param $\lambda$ & 0.1 \\
			Adam $\beta_1$ & 0\\
			Adam $\beta_2$ & 0.95\\
			Learning rate & $1e^{-2}$\\
			LR decay & 0.5\\
			L2 regularization & $3e^{-9}$\\
			\hline
		\end{tabular}
	\end{center}
	\caption{Hyper-parameters for baseline and high-order SRL models in our experiment. }
	\label{tab:hyper_both}
\end{table}

\begin{table}[h]
	\small
	\setlength{\tabcolsep}{2pt}
	\begin{center}
		\resizebox{\linewidth}{!}{
		\begin{tabular}{lcr}
			\hline \hline
			& \textbf{Version} & \textbf{Provider}\\ \hline
			CA & \textit{multi\_cased\_L-12\_H-768\_A-12} & \cite{devlin-etal-2019-bert} \\
			CS & \textit{Slavic BERT: slavic\_cased\_L-12\_H-768\_A-12} & \cite{arkhipov-etal-2019-tuning} \\
			DE & \textit{multi\_cased\_L-12\_H-768\_A-12} & \cite{devlin-etal-2019-bert} \\
			EN & \textit{wwm\_uncased\_L-24\_H-1024\_A-16} & \cite{devlin-etal-2019-bert} \\
			ES & \textit{BETO: spanish\_wwm\_cased\_L-12\_H-768\_A-12} & \cite{CaneteCFP2020} \\
			JA & \textit{NICT BERT: japanese\_100k\_L-12\_H-768\_A-12} &  \cite{nict-bert}\\
			ZH & \textit{chinese\_L-12\_H-768\_A-12} & \cite{devlin-etal-2019-bert} \\
			\hline
		\end{tabular}
	}
	\end{center}
	\caption{BERT versions for different languages. }
	\label{tab:bert_versions}
\end{table}

\subsection{Detail Results}\label{detailed_results}

The proverb that \textit{there is no such thing as a free lunch} tells us that no method works in every condition and scope. We explore our proposed high order structure learning for SRL in different languages and conditions: using pre-training or not, given or not given predicates, and different high-order structure combinations. We report all possible results on development sets, in-domain test sets, and out-of-domain test sets in Tables \ref{wp_all_results}, \ref{wp_all_results_1}, \ref{wop_all_results} and \ref{wop_all_results_1}. 
The experimental results illustrate the following points:

	1. In different languages, combinations of high-order structures bring different improvements. Some high-order structure combinations are even worse for performance in some languages.
	
	2. Pre-training can bring about a significant improvement in performance on both in-domain and out-of-domain test sets; however, the in-domain improvement is significantly greater than that of out-of-domain when the two domains are far apart. In particular, the difference between in-domain and out-of-domain in German and English is large, while the two domains in Czech are similar.
	
	3. The SRL results in German are lower than in other languages, the data analysis found that the proportion of predicates is very small, resulting in the sparse targets, which can not train the model well, especially when no predicates are pre-identified.

\begin{table*}[t!]
	\small
	\centering
	\begin{tabular}{llccccccccc}  
		\toprule  
		\multirow{2}{*}{Language} & \multirow{2}{*}{Method} &\multicolumn{3}{c}{Dev} &\multicolumn{3}{c}{Test} &\multicolumn{3}{c}{OOD} \\
		\cmidrule(lr){3-5}  \cmidrule(lr){6-8}  \cmidrule(lr){9-11}  & & P & R&F$_1$&P & R&F$_1$&P & R&F$_1$\\
		\midrule
		\multirow{16}{*}{CA} & baseline & 85.62 & 83.66 & 84.63 & 85.85 & 84.09 &	84.96 &  &  &   \\
		& +sib & 85.91 & 83.98 & 84.94 & 86.09 & 84.46 & 85.27 &  &  & \\
		& +cop & 85.78 & 83.83 & 84.79 & 85.97 & 84.17 & 85.06 &  &  & \\
		& +gp & 86.24 & 83.25 & 84.72 & 86.11 & 83.51 & 84.79 &  &  & \\
		& +sib+cop &  85.70 & 84.25 & 84.97 & 85.90 & 84.85 & \bf 85.37 &  &  &  \\
		& +sib+gp & 85.71 & 84.00 & 84.85 & 85.81 & 84.19 & 84.99 &  &  & 	\\
		& +cop+gp & 86.05 & 83.98 & \bf 85.00 & 85.74 & 83.72 & 84.72 &  &  &  \\
		& +\textbf{ALL} & 85.64 & 84.09 & 84.86 & 85.67 & 84.24 & 84.95 &  &  &  \\
		\cmidrule(lr){2-2} & baseline$^{\text{+B}}$ & 86.95 & 85.74 & 86.34 & 87.10 & 85.72 & 86.40 &  &  & \\
		& +sib &  86.94 & 85.92 & 86.43 & 87.15 & 86.05 & 86.59 &  &  &  \\
		& +cop & 87.15 & 85.96 & 86.55 & 87.14 & 86.00 & 86.57 &  &  & \\
		& +gp & 87.17 & 86.00 & 86.58 & 86.87 & 85.93 & 86.40 &  &  & \\
		& +sib+cop & 86.96 & 86.57 & 86.76 & 86.89 & 86.53 & 86.71 &  &  &   \\
		& +sib+gp & 87.26 & 86.55 & \bf 86.90 & 87.10 & 86.40 & 86.75 &  &  &  \\
		& +cop+gp & 87.24 & 86.12 & 86.68 & 87.08 & 85.97 & 86.52 &  &  &  \\
		& +\textbf{ALL} & 87.49 & 86.11 & 86.79 & 87.52 & 86.29 & \bf 86.90 &  &  &   \\
		\midrule
		\multirow{16}{*}{CS} & baseline & 91.20  & 89.79 & 90.49 & 90.87 & 89.49 &	90.18 & 91.22 & 89.88 &  90.54 \\
		& +sib &  91.30 & 89.71 & 90.50 & 91.08 & 89.56 & 90.32 & 91.25 & 89.72 & 90.48 \\
		& +cop &  91.40 & 89.77 & 90.58 & 91.09 & 89.56 & 90.32 & 91.27 & 89.75 & 90.51  \\
		& +gp &  91.35 & 89.63 & 90.48 & 91.16 & 89.40 & 90.27 & 91.31 & 89.62 & 90.46  \\
		& +sib+cop & 91.36 & 89.88 & \bf 90.81 & 91.33 & 89.89 & \bf 90.60 & 91.50 & 90.03 & \bf 90.75  \\
		& +sib+gp &  91.26 & 89.68 & 90.47 & 91.09 & 89.48 & 90.28 & 91.29 & 89.79 & 90.53 \\
		& +cop+gp & 91.24 & 89.61 & 90.42 & 91.01 & 89.51 & 90.26 & 91.23 & 89.73 & 90.48 \\
		& +\textbf{ALL} &  91.18 & 89.81 & 90.49 & 90.96 & 89.65 & 90.30 & 91.15 & 89.85 & 90.49  \\
		\cmidrule(lr){2-2} & baseline$^{\text{+B}}$ & 92.12 & 91.09 & 91.61 & 91.98 & 90.99 & 91.48 & 91.98 & 91.23 & 91.60 \\
		& +sib &  92.31 & 91.62 & \bf 91.96 & 91.98 & 91.23 & 91.60 & 91.94 & 91.50 & 91.72  \\
		& +cop & 92.11 & 91.27 & 91.69 & 92.08 & 91.25 & 91.66 & 91.97 & 91.49 & 91.73 \\
		& +gp & 92.02 & 91.19 & 91.60 & 91.97 & 91.20 & 91.58 & 91.85 & 91.32 & 91.59  \\
		& +sib+cop & 92.04 & 91.30 & 91.67 & 92.38 & 91.49 & \bf 91.93 & 91.84 & 91.47 & 91.65  \\
		& +sib+gp & 92.11 & 91.24 & 91.68 & 92.06 & 91.15 & 91.61 & 91.84 & 91.41 & 91.63 \\
		& +cop+gp & 91.99 & 91.32 & 91.65 & 91.94 & 91.25 & 91.60 & 91.87 & 91.61 & \bf 91.74 \\
		& +\textbf{ALL} & 92.03 & 91.26 & 91.65 & 91.99 & 91.14 & 91.56 & 91.75 & 91.32 & 91.53  \\
		\midrule
		\multirow{16}{*}{DE} & baseline & 75.83 & 72.51 & 74.13 & 77.48 & 74.61 &	76.02 &  71.34 & 67.73 &  69.49 \\
		& +sib & 76.63 & 73.36 & 74.96 & 77.01 & 75.54 & 76.27 & 71.66 & 69.36 & \bf 70.49  \\
		& +cop &  74.43 & 72.05 & 73.22 & 76.73 & 74.98 & 75.85 & 69.84 & 68.55 & 69.19  \\
		& +gp &  75.69 & 73.02 & 74.33 & 76.33 & 75.11 & 75.71 & 69.74 & 67.35 & 68.53 \\
		& +sib+cop & 76.24 & 73.25 & \bf 74.72 & 77.53 & 75.33 & \bf 76.41 & 71.28 & 68.88 & 70.06 \\
		& +sib+gp & 75.29 & 73.02 & 74.14 & 75.86 & 74.74 & 75.29 & 70.15 & 67.79 & 68.95 \\
		& +cop+gp &  76.22 & 72.05 & 74.08 & 77.00 & 74.25 & 75.60 & 69.99 & 66.76 & 68.33 \\
		& +\textbf{ALL} &  75.13 & 72.57 & 73.83 & 76.79 & 74.18 & 75.46 & 71.46 & 67.46 & 69.40 \\
		\cmidrule(lr){2-2} & baseline$^{\text{+B}}$ & 84.48 & 82.70 & 83.58 & 85.77 & 84.66 & 85.21 & 71.77 & 70.02 & 70.88 \\
		& +sib & 83.87 & 83.15 & 83.51 & 84.97 & 85.34 & 85.15 & 72.14 & 71.86 & 72.00 \\
		& +cop & 84.58 & 83.04 & 83.80 & 84.93 & 85.09 & 85.01 & 71.48 & 71.59 & 71.53 \\
		& +gp & 83.82 & 83.15 & 83.49 & 85.01 & 84.53 & 84.77 & 71.80 & 70.67 & 71.23 \\
		& +sib+cop & 84.67 & 83.61 & \bf 84.14 & 85.82 & 85.27 & \bf 85.54 & 72.23 & 71.48 & \bf 71.85 \\
		& +sib+gp & 84.46 & 82.58 & 83.51 & 85.21 & 84.10 & 84.65 & 71.93 & 69.04 & 70.45 \\
		& +cop+gp & 83.49 & 82.30 & 82.89 & 85.20 & 84.41 & 84.80 & 70.88 & 69.69 & 70.28 \\
		& +\textbf{ALL} & 84.53 & 82.70 & 83.60 & 84.95 & 84.17 & 84.56 & 71.39 & 68.17 & 69.74 \\
		\midrule
		\multirow{16}{*}{EN} & baseline & 90.15 & 86.27 & 88.17 & 91.29 &	88.00 &	89.61  & 81.37 & 77.12 & 79.19 \\
		& +sib & 89.94 & 86.67 & 88.27 & 91.46 & 88.53 & 89.97 & 81.76 & 77.85 & 79.76 \\
		& +cop & 90.03 & 86.59 & 88.27 & 91.40 & 88.45 & 89.90 & 81.89 & 78.07 & 79.94 \\
		& +gp & 89.86 & 86.13 & 87.96 & 91.41 & 88.14 & 89.75 & 82.20 & 78.51 & 80.31 \\
		& +sib+cop & 90.20 & 86.97 & \bf 88.55 & 91.33 & 88.74 & 90.02 & 81.80 & 78.36 & 80.04 \\
		& +sib+gp & 90.00 & 86.89 & 88.42 & 91.21 & 88.58 & 89.87 & 81.38 & 77.90 & 79.60 \\
		& +cop+gp & 89.69 & 86.58 & 88.11 & 91.26 & 88.68 & 89.95 & 81.32 & 78.02 & 79.64 \\
		& +\textbf{ALL} & 90.03 & 86.91 & 88.44 & 91.60 & 88.95 & \bf 90.26 & 82.6 & 78.75 & \bf 80.63 \\
		\cmidrule(lr){2-2} & baseline$^{\text{+B}}$ & 91.35 & 88.84 & 90.08 & 92.31 & 90.18 & 91.23 & 86.14 & 83.49 & 84.79 \\
		& +sib &  91.3	 & 89.11 & 	90.19	 & 92.49 & 	90.58 & 	91.53 & 	85.96 & 	83.97 & 	84.95 \\
		& +cop &  91.3	 & 89.19 & 	90.23 & 	92.21	 & 90.5 & 	91.35 & 	86.03 & 	84.07 & 	85.04 \\
		& +gp &   91.66	 & 88.6	 & 90.11 & 	92.77	 & 90.09	 & 91.41 & 	86.22 & 	83.1 & 	84.63 \\
		& +sib+cop &   91.16 & 	89.6	 & 90.37 & 	92.36 & 	90.84 & 	91.6	 & 85.59 & 	83.92	 & 84.75 \\
		& +sib+gp &  91.62 & 	88.99 & 	90.28 & 	92.64 & 	90.44 & 	91.53 & 	86.33 & 	83.61 & 	84.95 \\
		& +cop+gp & 91.27 & 	89.02 & 	90.14 & 	92.37 & 	90.35 & 	91.35 & 	86.03 & 	83.88 & 	84.94  \\
		& +\textbf{ALL} &  91.56 & 	89.35 & 	\bf 90.44	 & 92.59 & 	90.98 & 	\bf 91.77 & 	86.49 & 	83.80 & 	\bf 85.13  \\
		\bottomrule  
	\end{tabular}
	\caption{w/ pre-identified predicate results.}\label{wp_all_results}
\end{table*}

\begin{table*}[t!]
	\small
	\centering
	\begin{tabular}{llccccccccc}  
		\toprule  
		\multirow{2}{*}{Language} & \multirow{2}{*}{Method} &\multicolumn{3}{c}{Dev} &\multicolumn{3}{c}{Test} &\multicolumn{3}{c}{OOD} \\
		\cmidrule(lr){3-5}  \cmidrule(lr){6-8}  \cmidrule(lr){9-11}  & & P & R&F$_1$&P & R&F$_1$&P & R&F$_1$\\
		\midrule
		\multirow{16}{*}{ES} & baseline & 84.58 & 82.58 & 83.57 & 84.97 & 82.60 & 83.77 &  &  &    \\
		& +sib & 84.90 & 83.11 & \bf 83.99 & 85.18 & 83.28 & 84.22 &  &  &  \\
		& +cop & 84.66 & 82.76 & 83.70 & 85.12 & 83.21 & 84.15 &  &  & \\
		& +gp & 84.95 & 82.33 & 83.62 & 85.51 & 82.31 & 83.88 &  &  & \\
		& +sib+cop & 84.66 & 82.98 & 83.81 & 85.36 & 83.45 & \bf 84.39 &  &  &   \\
		& +sib+gp &  84.73 & 83.13 & 83.92 & 85.05 & 83.21 & 84.12 &  &  & \\
		& +cop+gp & 84.80 & 82.42 & 83.59 & 85.35 & 82.87 & 84.09 &  &  &  \\
		& +\textbf{ALL} & 84.89 & 82.90 & 83.89 & 85.12 & 83.29 & 84.20 &  &  & \\
		\cmidrule(lr){2-2} & baseline$^{\text{+B}}$ & 87.14 & 85.91 & 86.52 & 87.23 & 85.98 & 86.60 &  &  &  \\
		& +sib & 87.36 & 85.62 & 86.48 & 87.48 & 85.97 & 86.72 &  &  &   \\
		& +cop & 87.03 & 85.94 & 86.48 & 87.19 & 86.11 & 86.65 &  &  &  \\
		& +gp &  87.21 & 86.04 & 86.62 & 87.22 & 85.95 & 86.58 &  &  &  \\
		& +sib+cop & 86.98 & 86.45 & 86.71 & 87.24 & 86.67 & \bf 86.96 &  &  &  \\
		& +sib+gp & 87.62 & 85.66 & 86.63 & 87.62 & 85.84 & 86.72 &  &  &  \\
		& +cop+gp & 87.26 & 85.85 & 86.55 & 87.09 & 85.82 & 86.45 &  &  &  \\
		& +\textbf{ALL} & 87.49 & 86.11 & \bf 86.79 & 87.52 & 86.29 & 86.90 &  &  &  \\
		\midrule
		\multirow{16}{*}{JA} & baseline & 88.49 & 76.68 & 82.16 & 88.15 & 77.79 & 82.65 &  &  &  \\
		& +sib &  87.30 & 78.45 & 82.64 & 86.14 & 79.85 & 82.88 &  &  &  \\
		& +cop & 87.71 & 77.22 & 82.13 & 87.90 & 78.58 & 82.98 &  &  &  \\
		& +gp & 86.65 & 77.15 & 81.63 & 86.03 & 78.62 & 82.16 &  &  &  \\
		& +sib+cop &  87.97 & 78.78 & \bf 83.12 & 87.51 & 79.38 & \bf 83.25 &  &  & \\
		& +sib+gp & 88.32 & 77.72 & 82.68 & 88.34 & 78.51 & 83.14 &  &  &  \\
		& +cop+gp &  88.36 & 76.99 & 82.28 & 88.09 & 78.04 & 82.76 &  &  & \\
		& +\textbf{ALL} & 88.86 & 77.18 & 82.61 & 88.17 & 78.28 & 82.93 &  &  & \\
		\cmidrule(lr){2-2} & baseline$^{\text{+B}}$ & 89.93 & 80.89 & 85.17 & 89.63 & 81.83 & 85.55 &  &  &  \\
		& +sib &  89.29 & 81.06 & 84.98 & 89.02 & 82.08 & 85.41 &  &  &  \\
		& +cop & 89.99 & 80.56 & 85.02 & 89.71 & 81.57 & 85.45 &  &  &  \\
		& +gp & 89.43 & 80.40 & 84.67 & 88.88 & 81.08 & 84.80 &  &  &  \\
		& +sib+cop & 88.67 & 82.34 & \bf 85.39 & 88.65 & 83.32 & \bf 85.90 &  &  & \\
		& +sib+gp & 89.90 & 80.05 & 84.69 & 89.75 & 81.27 & 85.30 &  &  &  \\
		& +cop+gp &  89.20 & 81.27 & 85.05 & 88.59 & 82.03 & 85.19 &  &  &  \\
		& +\textbf{ALL} & 90.51 & 80.33 & 85.12 & 89.69 & 81.66 & 85.49 &  &  &  \\
		\midrule
		\multirow{16}{*}{ZH} & baseline & 87.28 & 83.84 & 85.52 & 87.95 & 83.63 & 85.73 &  &  &  \\
		& +sib &  87.58 & 83.96 & 85.73 & 87.94 & 83.80 & 85.82 &  &  &  \\
		& +cop & 88.33 & 83.40 & 85.80 & 88.61 & 83.29 & 85.87 &  &  &  \\
		& +gp & 88.08 & 82.25 & 85.07 & 88.46 & 82.01 & 85.12 &  &  &  \\
		& +sib+cop & 87.95 & 83.81 & 85.83 & 88.09 & 83.64 & 85.81 &  &  & \\
		& +sib+gp & 88.28 & 83.42 & 85.78 & 88.20 & 83.26 & 85.66 &  &  &  \\
		& +cop+gp & 88.38 & 82.56 & 85.37 & 88.54 & 82.63 & 85.48 &  &  &  \\
		& +\textbf{ALL} &  88.44 & 83.40 & \bf 85.85 & 88.35 & 83.82 & \bf 86.02 &  &  &  \\
		\cmidrule(lr){2-2} & baseline$^{\text{+B}}$ & 89.63 & 86.69 & 88.13 & 89.94 & 86.60 & 88.24 &  &  &  \\
		& +sib &  89.47 & 87.40 & 88.42 & 89.64 & 87.34 & 88.48 &  &  &  \\
		& +cop & 89.63 & 87.35 & \bf 88.48 & 89.79 & 87.33 & 88.54 &  &  &  \\
		& +gp & 89.16 & 86.74 & 87.93 & 89.54 & 86.69 & 88.09 &  &  &  \\
		& +sib+cop & 89.80 & 86.92 & 88.34 & 89.97 & 87.45 & \bf 88.69 &  &  &  \\
		& +sib+gp & 89.60 & 87.39 & \bf 88.48 & 89.79 & 87.30 & 88.53 &  &  & \\
		& +cop+gp & 89.48 & 87.18 & 88.32 & 89.70 & 87.36 & 88.52 &  &  &  \\
		& +\textbf{ALL} & 88.95 & 87.58 & 88.26 & 89.07 & 87.71 & 88.38 &  &  &  \\
		\bottomrule  
	\end{tabular}
	\caption{w/ pre-identified predicate results.}\label{wp_all_results_1}
\end{table*}

\begin{table*}[t!]
	\small
	\centering
	\begin{tabular}{llccccccccc}  
		\toprule  
		\multirow{2}{*}{Language} & \multirow{2}{*}{Method} &\multicolumn{3}{c}{Dev} &\multicolumn{3}{c}{Test} &\multicolumn{3}{c}{OOD} \\
		\cmidrule(lr){3-5}  \cmidrule(lr){6-8}  \cmidrule(lr){9-11}  & & P & R&F$_1$&P & R&F$_1$&P & R&F$_1$\\
		\midrule
		\multirow{16}{*}{CA} & baseline & 83.97 & 82.62 & 83.29 & 84.45 & 82.93 & 83.69 &  &  &   
		\\
		& +sib & 84.17 & 82.97 & 83.57 & 84.57 & 83.21 & 83.89 &  &  & 
		\\
		& +cop & 84.21 & 82.85 & 83.52 & 84.38 & 82.94 & 83.66 &  &  & 
		 \\
		& +gp & 84.52 & 82.23 & 83.36 & 84.55 & 82.28 & 83.40 &  &  & 
		\\
		& +sib+cop &  84.03 & 83.15 & 83.59 & 84.69 & 83.46 & \bf 84.07 &  &  & 
		 \\
		& +sib+gp & 84.15 & 83.06 & 83.60 & 84.35 & 83.07 & 83.70 &  &  & 
			\\
		& +cop+gp & 84.44 & 82.93 & \bf 83.68 & 84.32 & 82.64 & 83.47 &  &  & 
		  \\
		& +\textbf{ALL} & 83.95 & 83.07 & 83.51 & 84.14 & 83.10 & 83.62 &  &  & 
		  \\
		\cmidrule(lr){2-2} & baseline$^{\text{+B}}$ & 85.05 & 84.47 & 84.76 & 85.51 & 84.73 & 85.12 &  &  & 
		\\
		& +sib &  85.14 & 84.66 & 84.90 & 85.60 & 85.10 & 85.35 &  &  & 
		 \\
		& +cop & 85.46 & 84.80 & 85.13 & 85.57 & 84.99 & 85.28 &  &  & 
		\\
		& +gp & 85.33 & 84.81 & 85.07 & 85.29 & 84.93 & 85.11 &  &  & 
		 \\
		& +sib+cop & 85.39 & 85.27 & 85.33 & 85.47 & 85.38 & 85.42 &  &  & 
		  \\
		& +sib+gp & 85.39 & 85.27 & 85.33 & 85.47 & 85.38 & 85.42 &  &  & 
		 \\
		& +cop+gp & 85.39 & 84.89 & 85.14 & 85.53 & 85.05 & 85.29 &  &  & 
		  \\
		& +\textbf{ALL} & 86.08 & 85.35 & \bf 85.72 & 86.15 & 85.49 & \bf 85.82 &  &  & 
		 \\
		\midrule
		\multirow{16}{*}{CS} & baseline & 90.25 & 88.84 & 89.54 & 89.98 & 88.47 & 89.22 & 89.98 & 88.47 & 89.22
		 \\
		& +sib &  90.37 & 88.78 & 89.57 & 90.17 & 88.52 & 89.34 & 89.89 & 88.44 & 89.16
		 \\
		& +cop &  90.43 & 88.82 & 89.62 & 90.16 & 88.53 & 89.34 & 89.86 & 88.45 & 89.15
		  \\
		& +gp &  90.38 & 88.70 & 89.53 & 90.21 & 88.34 & 89.26 & 89.86 & 88.30 & 89.07
		  \\
		& +sib+cop & 90.73 & 88.94 & \bf 89.82 & 90.21 & 88.65 & 89.42 & 89.85 & 88.49 & 89.16
		  \\
		& +sib+gp &  90.34 & 88.77 & 89.55 & 90.15 & 88.41 & 89.27 & 89.88 & 88.51 & 89.19
		\\
		& +cop+gp & 90.32 & 88.72 & 89.51 & 90.12 & 88.47 & 89.29 & 89.86 & 88.44 & 89.14
		 \\
		& +\textbf{ALL} &  90.28 & 88.90 & 89.58 & 90.25 & 88.68 & \bf 89.45 & 89.82 & 88.85 & \bf 89.33
		  \\
		\cmidrule(lr){2-2} & baseline$^{\text{+B}}$ & 91.32 & 90.42 & 90.87 & 91.25 & 90.20 & 90.72 & 90.97 & 90.30 & 90.63
		 \\
		& +sib &  91.29 & 90.61 & 90.95 & 91.22 & 90.45 & 90.83 & 90.89 & 90.56 & 90.72
		 \\
		& +cop & 91.33 & 90.59 & \bf 90.96 & 91.32 & 90.50 & 90.91 & 90.99 & 90.58 & \bf 90.78
		 \\
		& +gp & 91.22 & 90.52 & 90.87 & 91.21 & 90.45 & 90.83 & 90.87 & 90.37 & 90.62
		  \\
		& +sib+cop & 91.25 & 90.63 & 90.94 & 91.23 & 91.21 & \bf 91.22 & 90.84 & 90.55 & 90.69
		 \\
		& +sib+gp & 91.31 & 90.55 & 90.93 & 91.31 & 90.37 & 90.84 & 90.82 & 90.48 & 90.65
		\\
		& +cop+gp & 91.18 & 90.63 & 90.91 & 91.20 & 90.47 & 90.83 & 90.84 & 90.66 & 90.75
		\\
		& +\textbf{ALL} & 91.22 & 90.55 & 90.89 & 91.23 & 90.37 & 90.80 & 90.74 & 90.42 & 90.58
		  \\
		\midrule
		\multirow{16}{*}{DE} & baseline & 53.59 & 68.87 & 60.27 & 51.03 & 72.95 & 60.06 & 39.97 & 45.14 & 42.40
		 \\
		& +sib & 53.81 & 69.89 & 60.81 & 51.32 & 73.63 & \bf 60.48 & 40.26 & 46.01 & 42.94
		  \\
		& +cop &  52.34 & 68.70 & 59.41 & 50.41 & 72.70 & 59.54 & 39.82 & 45.79 & 42.60
		 \\
		& +gp &  53.42 & 69.72 & 60.49 & 50.72 & 73.38 & 59.98 & 40.19 & 45.30 & 42.59
		 \\
		& +sib+cop & 53.28 & 70.29 & \bf 60.61 & 50.79 & 73.44 & 60.05 & 40.67 & 45.95 & \bf 43.15
		 \\
		& +sib+gp & 53.15 & 69.72 & 60.32 & 50.21 & 72.58 & 59.36 & 39.92 & 45.19 & 42.39
		 \\
		& +cop+gp &  53.13 & 68.53 & 59.86 & 51.17 & 72.77 & 60.09 & 40.14 & 44.87 & 42.37
		 \\
		& +\textbf{ALL} &  53.10 & 69.21 & 60.09 & 50.67 & 72.46 & 59.63 & 40.59 & 44.98 & 42.67
		\\
		\cmidrule(lr){2-2} & baseline$^{\text{+B}}$ & 57.87 & 80.14 & 67.21 & 55.67 & 83.18 & 66.70 & 37.99 & 43.13 & 40.40
		 \\
		& +sib & 57.68 & 80.76 & 67.30 & 55.70 & 83.98 & 66.98 & 39.00 & 44.38 & \bf 41.51
		 \\
		& +cop & 57.71 & 80.71 & 67.30 & 55.66 & 83.61 & 66.83 & 38.00 & 44.05 & 40.81
		 \\
		& +gp & 57.52 & 80.54 & 67.11 & 55.35 & 83.12 & 66.45 & 38.53 & 43.89 & 41.04
		 \\
		& +sib+cop & 58.00 & 81.33 & \bf 67.71 & 55.86 & 84.17 & \bf 67.15 & 38.94 & 44.38 & 41.48
		 \\
		& +sib+gp & 58.18 & 80.36 & 67.50 & 55.43 & 82.44 & 66.29 & 38.66 & 43.07 & 40.75
		 \\
		& +cop+gp & 57.23 & 79.97 & 66.71 & 55.43 & 82.99 & 66.47 & 38.34 & 42.86 & 40.47
		 \\
		& +\textbf{ALL} & 58.38 & 80.48 & 67.67 & 55.57 & 82.62 & 66.45 & 37.67 & 42.15 & 39.78
		 \\
		\midrule
		\multirow{16}{*}{EN} & baseline & 85.18 & 82.58 & 83.86 & 86.12 & 85.34 & 85.73 & 74.51 & 73.48 & 73.99
		 \\
		& +sib & 85.36 & 83.21 & \bf 84.27 & 86.00 & 85.64 & 85.82 & 74.38 & 73.31 & 73.84
		 \\
		& +cop & 85.31 & 82.91 & 84.09 & 86.12 & 85.56 & 85.84 & 74.35 & 73.07 & 73.70
		 \\
		& +gp & 85.40 & 82.67 & 84.01 & 86.04 & 85.13 & 85.58 & 74.43 & 72.68 & 73.55
		 \\
		& +sib+cop & 85.15 & 83.17 & 84.15 & 86.26 & 86.06 & \bf 86.16 & 74.76 & 73.65 & \bf 74.20
		\\
		& +sib+gp & 85.16 & 83.22 & 84.18 & 85.82 & 85.59 & 85.71 & 74.18 & 72.92 & 73.55
		 \\
		& +cop+gp & 84.93 & 83.07 & 83.99 & 86.00 & 85.59 & 85.79 & 74.29 & 73.31 & 73.80
		 \\
		& +\textbf{ALL} & 85.10 & 83.00 & 84.04 & 86.16 & 85.56 & 85.86 & 74.65 & 73.17 & 73.90
		 \\
		\cmidrule(lr){2-2} & baseline$^{\text{+B}}$ & 88.21 & 85.64 & 86.90 & 88.51 & 88.05 & 88.28 & 80.49 & 79.65 & 80.07
		 \\
		& +sib &  88.19 & 85.88 & 87.02 & 88.66 & 88.39 & 88.52 & 80.41 & 79.84 & 80.13
		 \\
		& +cop &  88.13 & 86.00 & 87.05 & 88.39 & 88.30 & 88.34 & 80.32 & 80.26 & \bf 80.29
		 \\
		& +gp &   88.54 & 85.41 & 86.95 & 89.01 & 87.98 & 88.49 & 80.63 & 79.24 & 79.93
		 \\
		& +sib+cop &   88.01 & 86.40 & \bf 87.20 & 88.55 & 88.60 & 88.57 & 79.87 & 79.89 & 79.88
		 \\
		& +sib+gp &  88.38 & 85.78 & 87.06 & 88.82 & 88.19 & 88.50 & 80.57 & 79.55 & 80.06
		 \\
		& +cop+gp & 88.16 & 85.79 & 86.96 & 88.59 & 88.20 & 88.40 & 80.35 & 80.11 & 80.23
		  \\
		& +\textbf{ALL} &  87.98 & 86.25 & 87.11 & 88.77 & 88.62 & \bf 88.70 & 80.01 & 79.80 & 79.90
		 \\
		\bottomrule  
	\end{tabular}
	\caption{w/o pre-identified predicate results.}\label{wop_all_results}
\end{table*}

\begin{table*}[t!]
	\small
	\centering
	\begin{tabular}{llccccccccc}  
		\toprule  
		\multirow{2}{*}{Language} & \multirow{2}{*}{Method} &\multicolumn{3}{c}{Dev} &\multicolumn{3}{c}{Test} &\multicolumn{3}{c}{OOD} \\
		\cmidrule(lr){3-5}  \cmidrule(lr){6-8}  \cmidrule(lr){9-11}  & & P & R&F$_1$&P & R&F$_1$&P & R&F$_1$\\
		\midrule
		\multirow{16}{*}{ES} & baseline & 83.35 & 81.79 & 82.57 & 83.52 & 81.58 & 82.54 &  &  &    
		\\
		& +sib & 83.72 & 82.42 & \bf 83.06 & 83.66 & 82.26 & 82.96 &  &  &  
		\\
		& +cop & 83.59 & 82.09 & 82.84 & 83.70 & 82.15 & 82.91 &  &  & 
		\\
		& +gp & 83.74 & 81.57 & 82.64 & 83.98 & 81.18 & 82.55 &  &  &  
		\\
		& +sib+cop & 83.41 & 82.11 & 82.75 & 83.87 & 82.38 & \bf 83.11 &  &  &   
		 \\
		& +sib+gp &  83.56 & 82.35 & 82.95 & 83.49 & 82.18 & 82.83 &  &  &  
		\\
		& +cop+gp & 83.65 & 81.61 & 82.62 & 83.87 & 81.86 & 82.85 &  &  & 
		\\
		& +\textbf{ALL} & 83.72 & 82.09 & 82.90 & 83.64 & 82.18 & 82.90 &  &  &  
		\\
		\cmidrule(lr){2-2} & baseline$^{\text{+B}}$ & 85.88 & 85.21 & 85.55 & 85.92 & 85.09 & 85.50 &  &  &  
		\\
		& +sib & 85.99 & 84.89 & 85.44 & 86.19 & 85.13 & 85.66 &  &  &   
		\\
		& +cop & 85.81 & 85.28 & 85.55 & 85.90 & 85.27 & 85.58 &  &  &   
		\\
		& +gp &  85.85 & 85.30 & 85.57 & 85.84 & 85.12 & 85.48 &  &  &   
		\\
		& +sib+cop & 85.66 & 85.64 & 85.65 & 86.21 & 85.75 & \bf 86.00 &  &  &  
		\\
		& +sib+gp & 86.20 & 84.93 & 85.56 & 86.32 & 85.02 & 85.66 &  &  &   
		\\
		& +cop+gp & 85.88 & 85.09 & 85.48 & 85.76 & 84.94 & 85.35 &  &  &  
		\\
		& +\textbf{ALL} & 86.08 & 85.35 & \bf 85.72 & 86.15 & 85.49 & 85.82 &  &  &   
		\\
		\midrule
		\multirow{16}{*}{JA} & baseline & 79.86 & 68.68 & 73.85 & 79.67 & 68.53 & 73.68 &  &  &   
		\\
		& +sib &  78.92 & 70.12 & 74.26 & 77.77 & 70.01 & 73.69 &  &  & 
		\\
		& +cop & 79.51 & 68.92 & 73.83 & 79.48 & 69.23 & 74.00 &  &  &
		\\
		& +gp & 79.23 & 69.39 & 73.98 & 78.14 & 69.24 & 73.42 &  &  & 
		\\
		& +sib+cop &  79.78 & 70.47 & \bf 74.84 & 79.14 & 69.57 & 74.05 &  &  &  
		\\
		& +sib+gp & 79.99 & 69.22 & 74.22 & 79.77 & 68.98 & 73.98 &  &  & 
		\\
		& +cop+gp &  80.45 & 69.04 & 74.31 & 79.91 & 69.01 & 74.06 &  &  & 
		\\
		& +\textbf{ALL} & 80.37 & 69.06 & 74.28 & 80.04 & 69.16 & \bf 74.20 &  &  & 
		 \\
		\cmidrule(lr){2-2} & baseline$^{\text{+B}}$ & 81.20 & 74.84 & 77.89  & 82.03 & 74.24 & 77.94  &  &  &  
		 \\
		& +sib &  82.77 & 74.49 & 78.41 & 80.53 & 76.86 & 78.65  &  &  & 
		\\
		& +cop & 82.61 & 74.35 & 78.27 & 82.33 & 75.36 & 78.69 &  &  & 
		\\
		& +gp & 82.15 & 74.16 & 77.95 & 80.98 & 75.48 & 78.13  &  &  & 
		\\
		& +sib+cop & 81.55 & 75.84 & \bf 78.59 & 82.16 & 75.85 & \bf 78.88  &  &  & 
		\\
		& +sib+gp & 82.28 & 73.62 & 77.71 & 82.10 & 75.04 & 78.41 &  &  & 
		\\
		& +cop+gp &  82.14 & 74.89 & 78.35 & 81.35 & 75.73 & 78.44 &  &  & 
		\\
		& +\textbf{ALL} & 82.72 & 74.21 & 78.23 & 81.88 & 75.39 & 78.50 &  &  & 
		 \\
		\midrule
		\multirow{16}{*}{ZH} & baseline & 82.00 & 80.20 & 81.09 & 83.08 & 79.91 & 81.46 &  &  & 
		\\
		& +sib &  82.34 & 80.42 & 81.37 & 83.04 & 80.00 & 81.50 &  &  & 
		 \\
		& +cop & 83.04 & 79.81 & 81.40 & 83.65 & 79.50 & 81.52 &  &  & 
		 \\
		& +gp &  82.84 & 78.73 & 80.73 & 83.71 & 78.25 & 80.89 &  &  &  
		\\
		& +sib+cop & 82.70 & 80.19 & 81.43 & 83.07 & 80.99 & \bf 82.01 &  &  & 
		\\
		& +sib+gp & 82.87 & 79.89 & 81.35 & 83.31 & 79.52 & 81.37 &  &  & 
		 \\
		& +cop+gp & 82.97 & 79.07 & 80.98 & 83.63 & 78.87 & 81.18 &  &  & 
		 \\
		& +\textbf{ALL} &  83.19 & 79.86 & \bf 81.49 & 83.51 & 79.59 & 81.50 &  &  & 
		 \\
		\cmidrule(lr){2-2} & baseline$^{\text{+B}}$ & 86.42 & 84.20 & 85.29 & 86.82 & 83.98 & 85.38 &  &  & 
		 \\
		& +sib &  86.24 & 84.78 & 85.51 & 86.47 & 84.63 & 85.54 &  &  & 
		\\
		& +cop & 86.34 & 84.72 & 85.52 & 86.55 & 84.54 & 85.53 &  &  &  
		\\
		& +gp & 85.90 & 84.24 & 85.07 & 86.37 & 84.00 & 85.17 &  &  & 
		\\
		& +sib+cop & 86.53 & 84.61 & \bf 85.56 & 86.83 & 84.57 & \bf 85.68 &  &  & 
		 \\
		& +sib+gp & 86.33 & 84.74 & 85.53 & 86.64 & 84.58 & 85.60 &  &  &  
		\\
		& +cop+gp & 86.28 & 84.63 & 85.45 & 86.61 & 84.66 & 85.63 &  &  & 
		\\
		& +\textbf{ALL} & 85.63 & 84.99 & 85.31 & 85.90 & 84.90 & 85.39 &  &  & 
		 \\
		\bottomrule  
	\end{tabular}
	\caption{w/o pre-identified predicate results.}\label{wop_all_results_1}
\end{table*}

\end{document}